\documentclass{article}

\usepackage{microtype}
\usepackage{graphicx}
\usepackage{subcaption}
\usepackage{booktabs} 

\usepackage{hyperref}

\usepackage{algpseudocode}
\usepackage{booktabs}

\usepackage[accepted]{icml_2026_arxiv}

\usepackage{amsmath}
\usepackage{amssymb}
\usepackage{mathtools}
\usepackage{amsthm}
\usepackage{xcolor}
\usepackage{listings}

\usepackage[capitalize,noabbrev]{cleveref}

\theoremstyle{plain}

\theoremstyle{definition}

\theoremstyle{remark}

\usepackage[textsize=tiny]{todonotes}

\icmltitlerunning{Structuring The Future: Diffusion LLM Speculative Decoding via Calibrated Draft Graphs}

\begin{document}

\twocolumn[
  \icmltitle{Structuring The Future: \\ Diffusion LLM Speculative Decoding via Calibrated Draft Graphs}



  \icmlsetsymbol{equal}{*}

  \begin{icmlauthorlist}
    \icmlauthor{Sudhanshu Agrawal}{QAIR}
    \icmlauthor{Risheek Garrepalli}{QAIR}
    \icmlauthor{Raghavv Goel}{QAIR}
    \icmlauthor{Christopher Lott}{QAIR}
    \icmlauthor{Fatih Porikli}{QAIR}
    \icmlauthor{Mingu Lee}{QAIR}
  \end{icmlauthorlist}

  \icmlaffiliation{QAIR}{Qualcomm AI Research}

  \icmlcorrespondingauthor{Sudhanshu Agrawal}{sudhagra@qti.qualcomm.com}

  \icmlkeywords{Machine Learning, ICML}

  \vskip 0.3in
]


\printAffiliationsAndNotice{}  

\begin{abstract}
Diffusion LLMs (dLLMs) have recently emerged as a powerful alternative to autoregressive LLMs (AR-LLMs) with the potential to operate at significantly higher token-generation rates. To unlock this potential, we present Spiffy, a speculative decoding algorithm to accelerate dLLM inference while provably preserving the model's output distribution. This work addresses the unique challenges involved in applying ideas from speculative decoding of AR-LLMs to dLLMs. Spiffy performs auto-speculation to eliminate the overheads of an independent draft model, structuring draft states in the form of a novel directed draft graph to take advantage of the bidirectional, blockwise nature of dLLM generation. These draft graphs are calibrated offline to maximize acceptance rates and are dynamically pruned during inference for improved computational efficiency. We present a detailed formulation of Spiffy and demonstrate its ability to accelerate LLaDA, Dream, and SDAR models in combination with KV caching and threshold-based dynamic unmasking leading to up to $8.6\times$ reduction in model inferences and $6.3\times$ acceleration in token rate. 
\end{abstract}

\section{Introduction}

The majority of large language models (LLMs) today are autoregressive LLMs (AR-LLMs) \cite{brown2020language}. Such models operate left-to-right, generating new tokens conditioned on the previous tokens in the sequence. AR-LLMs have demonstrated impressive capabilities in a variety of domains but, due to their autoregressive nature, are constrained to generating a single token per model inference. As overall latency is dominated by the total model inference time, this constraint is a bottleneck in LLM decoding speeds.

More recently, the emergence of diffusion LLMs (dLLMs) \cite{llada, mercury} represents a paradigm shift in language modeling. dLLMs are inherently parallel in nature and are not bound by a strict causal factorization. Instead, they model the joint distribution over token blocks, leveraging bidirectional attention \cite{bert}, and enabling arbitrary-order multi-token decoding. While commercial dLLMs such as Mercury \cite{mercury}, Gemini Diffusion \cite{Google_2025}, and Seed Diffusion \cite{song2025seeddiffusionlargescalediffusion} may achieve token rates in excess of 1,000 tokens/second, open-sourced dLLMs such as LLaDA \cite{llada} and Dream \cite{dream2025} do not yet achieve these decoding speeds. In particular, to maintain output quality, these dLLMs generate only a \textit{single token} per model inference by default, restricting their efficiency. Recent works seek to improve dLLM inference efficiency via KV caching \cite{fastdllmtrainingfreeaccelerationdiffusion, ma2025dkv} and threshold-based dynamic unmasking \cite{fastdllmtrainingfreeaccelerationdiffusion}.

On the other hand, recent work on AR-LLMs circumvents their autoregressive constraints via speculative decoding \cite{leviathan2023fast, chen2023acceleratinglargelanguagemodel}. This has emerged as a popular and powerful method to accelerate AR-LLMs due to its accuracy guarantees and potential for efficient deployment. However, transferring the ideas developed for AR-LLM speculation to dLLMs encounters fundamental challenges. Since dLLM token distributions do not yield a causal factorization, draft verification must be adapted to remain lossless. Typical drafting strategies, such as tree-based drafting, also require substantial rethinking. Further, the choice of a compatible and efficient drafter for dLLMs is essential. In this work, we address these challenges and develop speculative decoding for dLLMs.

\begin{figure}[h]
    \centering
    \includegraphics[width=0.70\columnwidth]{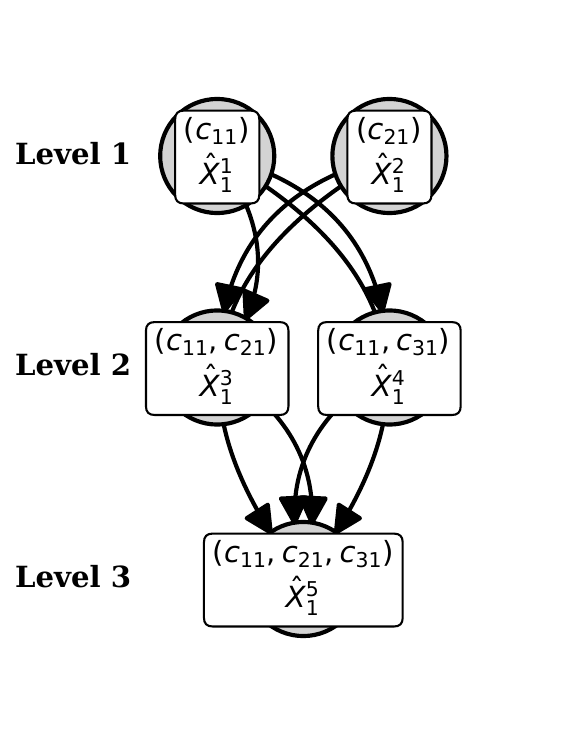}
    \caption{Spiffy defines and calibrates directed draft graph structures to capture multi-token unmasking dynamics. Spiffy's draft graphs leverage the bidirectional nature of dLLM decoding, allowing a draft state to be accepted via multiple potential pathways.}
    \label{fig:sample_graph_viz}
\end{figure}

\textbf{Spiffy} (\textbf{Sp}eculation for D\textbf{iff}usion LLM Efficienc\textbf{y}), is a novel speculative decoding algorithm for dLLMs. During generation, Spiffy samples draft states from the target dLLM distribution itself, avoiding the need for, and costs of an auxiliary draft model. Spiffy structures these drafts as a directed draft graph to maximize their acceptance. While these draft graphs are similar in motivation to draft trees \cite{li2024eagle, li2025eagle3scalinginferenceacceleration, miao2024specinfer, jeon2024recursive}, Spiffy's draft graphs respect and also take advantage of the bidirectional nature of dLLM generation. To optimize these graph structures, we introduce a novel offline calibration procedure which determines effective draft graph configurations. These calibrated graphs significantly increase acceptance rates during generation and are dynamically pruned during inference for computational efficiency. In this work, we define speculative decoding for dLLMs and demonstrate our algorithm's ability to unlock accelerated dLLM inference. We summarize our key contributions below. 

\subsection*{Summary of Contributions}
\begin{enumerate}
    \item \textbf{Speculative decoding of dLLMs:} We introduce Spiffy, a novel speculative decoding algorithm. We define drafting and verification for dLLMs as seamless, inexpensive procedures, and prove Spiffy's lossless-acceleration property formally and experimentally.
    \item \textbf{Directed draft graphs:} We define directed draft graphs, a generalization of speculative draft trees, designed to take advantage of the bidirectional nature of dLLMs and maximize draft acceptance. 
    \item \textbf{Graph structure calibration and pruning:} We develop a novel calibration algorithm to generate optimized draft graphs that capture token-unmasking dynamics, boosting acceptance rates during speculation. During inference, these draft graphs are dynamically pruned to improve computational efficiency. 
    \item \textbf{Auto-speculation:} Spiffy leverages the target dLLM distribution at every timestep to speculate draft states, thus avoiding the need to train and run a draft model. 
    \item \textbf{Experimental validation:} We validate our methods through comprehensive experimentation with multiple open-source dLLMs (LLaDA-8B-Instruct, Dream-7B-Instruct, SDAR-8B-Chat-b32), evaluated on standard benchmarks (GSM8K, HumanEval, MATH500, MBPP), in combination with prefix KV caching and threshold-based dynamic unmasking, enabling up to $8.6\times$ reduction in model inferences and $6.3\times$ speedups in token rate. 
\end{enumerate}

\section{Related Work}
\label{sec:related}
\subsection{Diffusion LLMs} 
Diffusion models \cite{pmlr-v37-sohl-dickstein15, ho2020denoising} have become a dominant approach in generative modeling, particularly excelling in continuous domains, such as image synthesis \cite{song2020score, saharia2022photorealistic, esser2024scaling, labs2025flux}. Their capacity for parallel generation has motivated their adaptation to language modeling, a discrete domain. Early works explored structured denoising and discrete diffusion processes for text generation \cite{austin2021structured, lou2023discrete}, while more recent efforts have focused on token-level masking-based forward processes \cite{sahoo2024simple, shi2024simplified} and blockwise decoding \cite{arriola2025block, cheng2025sdarsynergisticdiffusionautoregressionparadigm}. Current open-source models implementing these ideas at scale include LLaDA \cite{llada, bie2025llada2}, Dream \cite{dream2025,xie2025dream}, SDAR \cite{cheng2025sdarsynergisticdiffusionautoregressionparadigm}, DiffuCoder \cite{gong2025diffucoder}, and OpenMoE \cite{ni2025openmoe2}. 

\subsection{Accelerating Diffusion LLMs} 
While dLLMs have the potential for parallel generation, in practice, many still decode one token at a time, thus requiring a large number of denoising steps. To overcome this, recent work has explored simultaneous token updates based on notions of confidence determined intrinsically \cite{fastdllmtrainingfreeaccelerationdiffusion} or by an auxiliary AR-LLM \cite{israel2025acceleratingdiffusionllmsadaptive}, as well as the use of modified solvers to accelerate the reverse diffusion process \cite{shaul2024flow, li2025lavida, luxembourg2025plan}. Recent work also proposes drafting and verification for dLLMs \cite{hong2025wide, gao2025selfspeculativedecodingdiffusion}, but without the use of structured directed draft graphs. Additionally, efforts have been made to integrate KV caching mechanisms to reduce redundant computation \cite{ma2025dkv, liu2025dllm, liu2025dllmcacheacceleratingdiffusionlarge}. An alternate line of work focuses on improving efficiency through step distillation \cite{deschenaux2025beyond, qian2026d3llm}. 

\subsection{Speculative Decoding of AR-LLMs} Speculative decoding \cite{leviathan2023fast, chen2023acceleratinglargelanguagemodel} accelerates the inference of a large, target model with an efficient drafter. Generations from the drafter have seen several forms, most often tree-based \cite{miao2024specinfer, jeon2024recursive, li2024eagle}. Other works reduce the overhead of the drafter \cite{lin2025bita} or have tried to do away with it entirely \cite{lookaheadfu2024break, zhang2023draft, fu2024break}. Recent methods have proposed improvements to the draft generation quality via training improvements \cite{goel2024direct}, by allowing the drafter to leverage the final hidden features of the target model \cite{li2024eagle}, and by adding additional linear modules to predict multiple draft tokens at once \cite{cai2024medusa}. Recent works also explore the use of a small diffusion language model as a draft model for a target AR-LLM \cite{christopher2024speculative}.

\subsection{Speculation for Image Diffusion Models} \cite{wang2024continuous} and \cite{de2025accelerated} concurrently propose modifications to rejection sampling in the case of continuous-valued diffusion models and achieve impressive speedups in image generation latency. While \citeauthor{wang2024continuous} use an independent diffusion model as a draft model, \citeauthor{de2025accelerated} and \citeauthor{hu2025diffusion} use the target model to speculate future states. Other works explore the use of multi-sample generation and gradient-based MCMC within energy-based formulations \cite{du2020compositional, singhal2025general}. 

\begin{figure*}[t]
    \centering
    \includegraphics[width=0.8\textwidth]{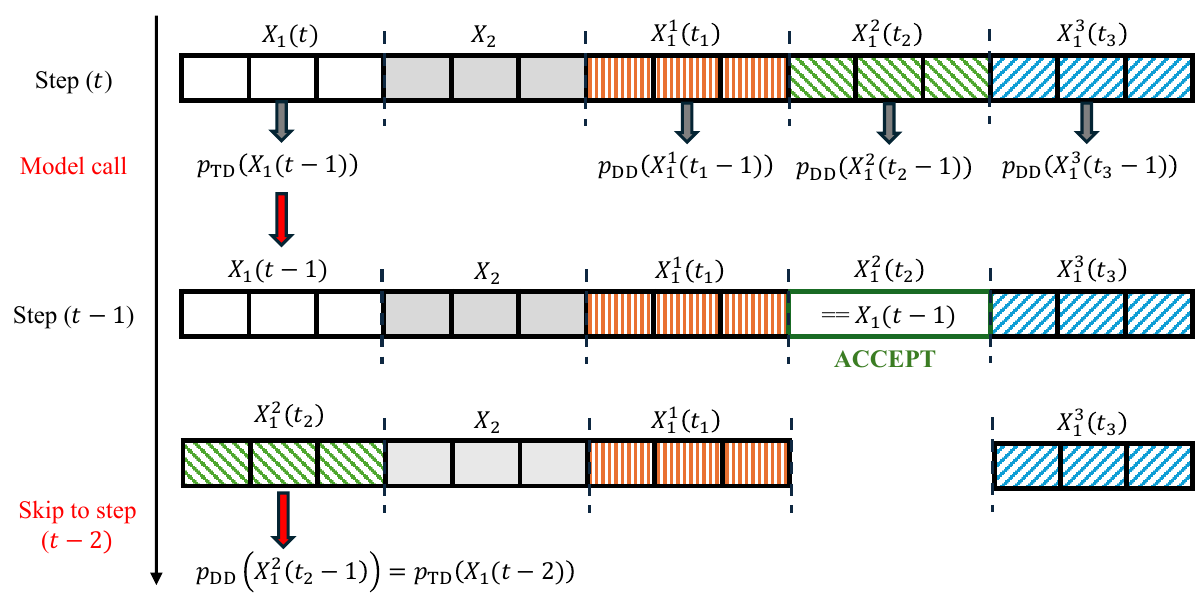}
    \caption{Spiffy's lossless verification allows $X_1(t)$ to advance to state $X_1(t-2)$ using a single model inference. Draft blocks $X_1^1, X_1^2, X_1^3$ are appended to the sequence, and model inference is performed in parallel. Each draft is compared against $X_1(t-1)$ to check for acceptance. If a draft is accepted, we use its predicted draft distribution to skip ahead and predict $X_1(t-2)$. We then repeat this process with the remaining drafts to check for additional acceptances.}
    \label{fig:verification}
\end{figure*}

\section{Preliminaries}
\label{sec:prelim}
\subsection{Masked Diffusion Language Models} 
Masked diffusion language models, at inference time, define a reverse denoising process where a token sequence $X(T)$ is iteratively unmasked to a valid sequence $X(0)$. Practically, this is implemented by setting a generation length $W$ and initializing the sequence with a specially designated \texttt{MASK} token as $X(T) = [\texttt{MASK}, \cdots, \texttt{MASK}]$. This sequence is divided into $N$ blocks of size $L$ each, and we denote the $i^{th}$ block of $X$ by $X_i$. The dLLM then fully unmasks each block one by one, from $X_i(T) \rightarrow X_i(0)$. In particular, at each timestep $t+1\in \{T, \cdots, 0\}$, the dLLM models:
\begin{align}
    \label{eq:dllm}
    p_\theta\left(X_i(t)\mid X(t+1)\right)
\end{align}
By sampling $X_i(t)$ from \eqref{eq:dllm}, the next candidate state of the sequence with per-token probabilities $p_1, \cdots, p_L$ is obtained. We define the rate of unmasking as $S_t$, and the top-${S_t}$ tokens, according to their probabilities, are unmasked while the others remain masked. $S_t$ thus represents the rate at which a block is denoised. Open-source dLLMs such as LLaDA \cite{llada} and Dream \cite{dream2025} often set this parameter as $S_t = S = 1$ for all timesteps $t$, consequently unmasking one token per iteration. This setting allows these models to achieve accuracies comparable to AR-LLMs. 

Recently, \textbf{threshold-based dynamic unmasking} \cite{fastdllmtrainingfreeaccelerationdiffusion} has proved to be an effective solution to this issue. A fixed threshold $\tau$ is determined, and at each iteration, all tokens with probability $p_i > \tau$ are unmasked, leading to a variable unmasking rate $S_t$ per iteration. \citeauthor{fastdllmtrainingfreeaccelerationdiffusion} show that this method preserves accuracy while speeding up dLLM generations significantly. Other recent work involves KV caching of dLLMs via \textbf{prefix-caching} \cite{fastdllmtrainingfreeaccelerationdiffusion, ma2025dkv} where the KVs of all $X_{<i}$ remain fixed while decoding $X_i$. We present the speedup of our speculative decoding algorithm in combination with these methods and demonstrate its ability to improve upon their acceleration.

\subsection{Speculative Decoding}
Speculative decoding \cite{leviathan2023fast, chen2023acceleratinglargelanguagemodel} is a powerful technique designed to parallelize the decoding process of AR-LLMs while provably preserving their output distribution. A typical speculative decoding setup consists of a large \textit{target} model that one wishes to accelerate and a second, more efficient \textit{drafter} that approximates it. This drafter may take many forms, as discussed in Sec. \ref{sec:related}, and proposes a set of draft completions for the target model to verify in parallel. Concretely, given a prompt $X_P = [x_1, \cdots, x_t]$, the target model has a target distribution $p_{\text{TD}}$ and autoregressively models $p_{\text{TD}}\left(x_i \mid x_1, \cdots, x_{i-1}\right)$ for $i\in \{t+1,\cdots, t+L\}$, thus generating $L$ new tokens, $\left[x_{t+1}, \cdots, x_{t+L}\right]$. This process is slow, so, to accelerate it, the drafter models a draft distribution $p_{\text{DD}}\left(x_i \mid x_{<i}\right)$ generating draft tokens $[\hat{x}_{t+1}, \cdots, \hat{x}_{t+D}]$ for some draft length $D$. These draft tokens may then be \textit{verified} in parallel by obtaining $p_{\text{TD}}(\cdot\mid x_{<i}, \hat{x}_{<i})$ and performing rejection sampling with the candidate draft tokens. In a greedy-decoding setting, we may simply sample $x_i \sim p_{\text{TD}}(\cdot \mid x_{<i}, \hat{x}_{<i})$ and accept $\hat{x}_i$ if it equals $x_i$. If we accept $M$ of these draft tokens, we have effectively sampled $\left[x_{t+1}, \cdots, x_{t+M}\right]$ using only a single target model inference. 

\section{Method}

Spiffy casts the denoising process of dLLMs in a speculative decoding framework.  Rather than speculating on individual tokens, Spiffy speculates the state of the sequence over denoising timesteps, allowing us to {\it skip ahead} in the unmasking process while maintaining output quality.

\subsection{Notation} Suppose, in the setting defined in Sec. \ref{sec:prelim}, we are currently denoising block $k\in \{1, \cdots, N\}$ and $X_k$ is observed at a denoising timestep $t \in [T, 0]$ as $X_k(t)$. Since the rest of the sequence is unchanged while $X_k$ is being unmasked, we refer to the rest of the sequence collectively as $X'$. Thus, the token sequence as a whole is $X(t) = X';X_k(t)$ (the sequence with $X_k(t)$ taking the position of the $k^{th}$ block).

\subsection{Draft Blocks}
When unmasking any given token, it is essential to consider the influence of every other token in the sequence. This is a characteristic property of bidirectional attention \cite{bert} and to satisfy this condition, we consider drafts at the block level. We denote a \textbf{draft block} for $X_k$ as $\hat{X}_k^m(t_m)$ with $m \in\{1, \cdots, D\}$ for a number of drafts, $D$. Each draft block represents a speculation of the state of the block at a timestep $t_m$ and we refer to the set of current draft blocks as $\{\hat{X}_k^m\}$. 

\subsection{Verification}
\label{sec:verification}

\begin{figure}[ht]
    \centering
    \includegraphics[width=0.7\columnwidth]{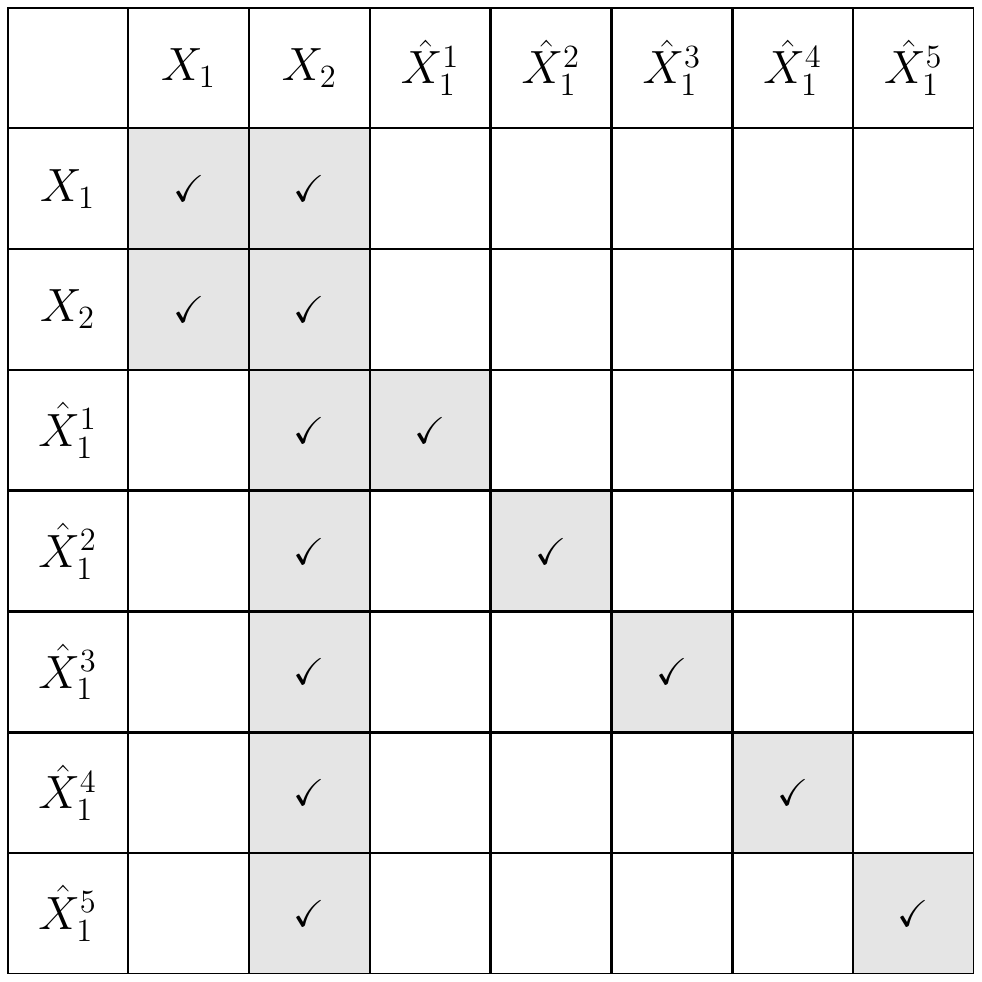}
    \caption{Spiffy constructs a blockwise attention mask to verify the draft candidate states. The true sequence continues to attend to itself. Each draft block attends to itself and all blocks in the true sequence except the original block for which it is drafting. For blockwise-causal dLLMs, this mask is updated to have a blockwise-causal masking structure with respect to the true tokens.}
    \label{fig:attn-mask}
\end{figure}

At timestep $t$, the current state of the target (true) sequence is $X';X_k(t)$. We use the dLLM to model two separate distributions, the target distribution at the next timestep: 
\begin{align}
    p_{\text{TD}}(t-1)=p_\theta\left(X_k(t-1)\mid X';X_k(t)\right)
\end{align}
and the draft distributions at their corresponding timesteps: 
\begin{align}
    p_{\text{DD}}^m(t_m-1) = p_\theta\left(X_k(t_m-1)\mid X';\hat{X}_k^m\right)
\end{align}
This can be achieved in parallel using a single model call with a custom attention mask and appropriate positional embeddings, similar to the strategies adopted by \cite{miao2024specinfer,jeon2024recursive, li2024eagle}. We include an example of such a mask in Fig. \ref{fig:attn-mask}. 

To advance the sequence to the next timestep, $S_{t-1}$ tokens are sampled from the target distribution and are unmasked. The state of the block is then compared to the draft candidates and any draft that matches is accepted. Since we already have access to the draft distribution of the accepted block, we may skip ahead in the denoising and repeat the procedure with the remaining drafts. This verification method is described in detail in Algorithm \ref{alg:verification} and visualized in Fig. \ref{fig:verification}. 

\begin{algorithm}[h]
\caption{Spiffy Draft Block Verification}
\label{alg:verification}
\begin{algorithmic}
\State \textbf{Given:} unmasking rate $\{S_t\}$, vocabulary of size $V$
\State \textbf{Input:} current $t$, current $X_k(t)$, target $p_{\text{TD}}$, draft $\{p_{\text{DD}}^m\}$, drafts $\{\hat{X}_k^m\}$

\Statex \textbf{(1) Extract marginals $p_n$ for each token position}
\State $p_n \gets p_{\text{TD}}(t-1)[n] \in [0, 1]^{V}$ 
\Statex \textbf{(2) top-1 probability for each token position $n$}
\State $p^1_n, j_n \gets \max, \arg\max_{v \in V} p_n(v)$  $\forall\; n\in [1, L]$
\Statex \textbf{(3) Unmask top-$S_{t-1}$ positions}
\State $n_1, \dots, n_{S_{t-1}} \gets \text{topKIndex}(S_{t-1}, [p^1_1, \dots, p^1_L])$ 
\State $c_{n_i} \gets \text{vocabulary}[j_{n_i}] \quad \forall i\in 1, \cdots, S_{t-1}$ 
\Statex \textbf{(4) Update the block with the new tokens}
\State $X_k(t-1) \gets X_k(t)$
\State $X_k(t-1)[n_i] \gets c_{n_i}\quad \forall\; i\in 1, \cdots, S_{t-1}$ 
\State $t \gets t-1$ 
\smallskip
\Statex \textbf{(5) Draft verification loop}
\For{$m \in \{1, \dots, D\}$} 
    \If{$t_m \neq t$} 
        \State \textbf{continue}
    \EndIf
    \If{$\hat{X}_k^m = X_k(t)$} 
        \Statex \hspace{2.75em} \textbf{\# Accept}
        \State $p_{\text{TD}}(t-1) \gets p_{\text{DD}}^m(t-1)$ 
        \State Pop $\hat{X}_k^m$ from the draft list
        \Statex \hspace{2.75em} \textbf{\# Restart the algorithm from this state}
        \Statex \hspace{3.5em} \textbf{using $children(\hat{X}_k^m)$ \eqref{eq:draft-graph-1} \eqref{eq:draft-graph-2} \eqref{eq:draft-graph-3}} 

        \State \textsc{Verify}$(t, X_k(t), p_{\text{TD}}$, 
        \Statex \hspace{4em} $\{p_{\text{DD}}\}_{\in children( \hat{X}_k^m)}, \{\hat{X}_k\}_{\in children (\hat{X}_k^m)})$
    \EndIf
\EndFor

\smallskip
\State \textbf{Output: } $t$, $X_k(t)$
\end{algorithmic}
\end{algorithm}

Thus, if in the process of unmasking a block at a rate of $S_t$ tokens, we accept drafts for timesteps $t_1, \cdots, t_M$, we decrease the number of model inferences required by a factor of $(T-M)/T$. 

We show that Algorithm \ref{alg:verification} is \textit{lossless}, that is, it preserves the output distribution of the model. We include a proof of this statement in Appendix \ref{appendix:proof-formal}. In certain settings, producing the exact draft probabilities $\{p^m_{\text{DD}}\}$ is computationally expensive, and achieving exact decoding requires construction of a batched attention mask and batched token sequence which would proportionally increase the computational requirements of the system. Instead, we choose to create a blockwise tree-attention mask structure that achieves near-lossless performance for full-sequence bidirectional dLLMs such as LLaDA. Moreover, using this simple mask, we can achieve exact lossless speculative decoding when utilizing dual caches \cite{fastdllmtrainingfreeaccelerationdiffusion} or when decoding blockwise causal dLLMs such as SDAR \cite{cheng2025sdarsynergisticdiffusionautoregressionparadigm}. Further discussion, including the impact of floating-point precision loss, is included in Appendix \ref{appendix:metrics}. 

\subsection{Drafting}
\label{sec:drafting}

\subsubsection{Draft Graphs}
\label{sec:draft-graphs}
We now desire draft blocks $\{\hat{X}_k^m\}$ for $m\in \{1, \cdots, D\}$ such that their \textit{acceptance is maximized}. In particular, we would like the opportunity to have \textit{multiple} drafts accepted per model call. To this end, we structure the draft blocks to have a directed, \textit{parent-child} relationship with each other and to the current block $X_k(t)$. Block $A$ at timestep $t_a$ is defined to be a \textbf{parent} of block $B$ at timestep $t_b$ if:
\begin{align}
\label{eq:draft-graph-1}
t_b &= t_a - 1 \\
\label{eq:draft-graph-2}
|unmasked(A)| + S_{t_b} &= |unmasked(B)| \\
\label{eq:draft-graph-3}
unmasked(A) &\subset unmasked(B)
\end{align}
Intuitively, block $B$ should be a potential \textit{next step} in the denoising process starting from block $A$. This property allows us to accept multiple drafts per model call since the acceptance of draft $A$ means that we may potentially accept any children of $A$ that were also drafted. In this manner, we may skip ahead multiple timesteps with a single model call. 

We define $children(A)$ as the set of all blocks $B$ that satisfy (\ref{eq:draft-graph-1}, \ref{eq:draft-graph-2}, \ref{eq:draft-graph-3}). We first choose a set of draft blocks,  $\{\hat{X}_k^m\}(\text{level }1) \subset children(X_k(t))$, representing the \textit{first level} of drafts. Draft blocks for subsequent levels are then chosen in a similar manner:  
\begin{align}
    \{\hat{X}_k^m\}(\text{level }i) \quad \subset \bigcup_{A\in \{\hat{X}_k^m\}(\text{level } i-1)} children(A)
\end{align}

Together, our draft blocks $\{\hat{X}^m_k \}$ take on the form of a \textbf{directed draft graph}. Notably, unlike draft trees used for speculative decoding of AR-LLMs, each draft block may have \textit{multiple parents} and thus \textit{multiple pathways to being accepted}, a unique advantage of bidirectional attention. A visualization of such a draft graph is provided in Fig. \ref{fig:sample_graph_viz}.

\subsubsection{Drafting Source} Given this desired drafting structure, we now need a process by which the contents, i.e., unmasked tokens, of a draft block may be determined. At any timestep, we have access to $p_{\text{TD}}=p_\theta\left(X_k(t-1)\mid X';X_k(t)\right)$, which we leverage to construct draft blocks. For brevity, ArgSort is assumed to be in descending order in the following relations.

First, with the marginals $p_1, \cdots, p_L$ of the current distribution, we obtain an ordering of token positions: 
\begin{align}
    \label{eq:token-position-sort}
    &n_1, \dots, n_{L} \gets \text{ArgSort}\left(\max\left(p_1\right), \dots, \max\left(p_L\right)\right) 
\end{align}

Second, for each sorted position $n_i$, we determine the top-$k$ token choices from the vocabulary of size $V$: 
\begin{align}
    \label{eq:vocab-sort}
    &c_{i1}, \cdots, c_{ik} \gets \text{ArgSort}(p_{n_i}(v_1), \cdots, p_{n_i}(v_V))[\:\text{: }k\;] 
\end{align}

\begin{figure}[h]
    \centering
    \makebox[\linewidth][c]{\hspace*{-0.6cm}\includegraphics[width=0.70\columnwidth]{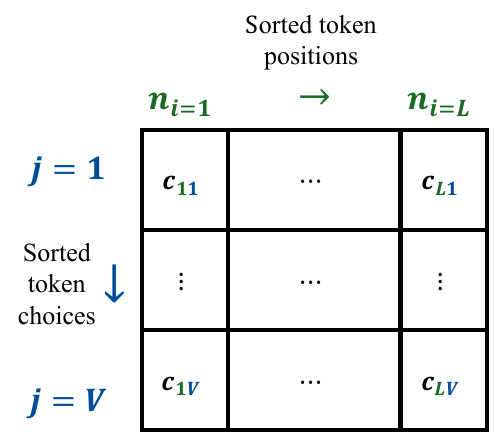}}
    \caption{By leveraging the marginals of $p_{\text{TM}}\left(X(t)\right)$ for each position in the block, we sort token candidates by $i=$ token position rank and $j=$ vocabulary rank.}
    \label{fig:indices_viz}
\end{figure}

Thus, when considering candidates for unmasking, the `ranking' of a candidate depends on both its position in the token sequence and its likelihood among token choices for that position. We can characterize this ranking by defining two new terms. For any token position $n\in \{1, \cdots, L\}$ that we are considering unmasking, we define its \textbf{token position rank} and denote it by $i$
\begin{align}
\label{eq:i-select}
    &i := \text{ArgSort}\left(\max(p_1), \cdots, \max(p_L)\right)[\: n\: ]  
\end{align}
Further, for token position $n$, for any particular token $v\in \{1, \cdots, V\}$ that we could unmask in this position, we define the \textbf{vocabulary rank} of this choice $v$ and denote it by $j$
\begin{align}
\label{eq:j-select}
    &j := \text{ArgSort}\left(p_n(1), \cdots, p_n(V)\right)[\: v \:] 
\end{align}
So any candidate token that may be unmasked at a sequence position can be denoted by $c_{ij}$ in terms of its token position rank and its vocabulary rank. These indices $(i, j)$ and the sorting method is visualized in Fig. \ref{fig:indices_viz}. In this way, we define each of our draft blocks $\hat{X}_k^m (t_m)$ as consisting of the current block state $X_k(t)$ with a set of additionally unmasked tokens $\{c_{ij}\}$, such that $\left\lvert\{c_{ij}\}\right\rvert = \sum_{t<t'\leq t_m}S_{t'}$. By leveraging the current target distribution in the construction of draft blocks, we avoid the need for an independent draft model. 

\subsubsection{Draft Graph Calibration}
Since we now have a source for draft blocks, the next step is to determine which candidate states are \textit{most promising}. We know that it is to our advantage to structure these drafts as a directed graph, but since there are $L$ possible unmasking positions, each with $V$ candidates to choose from, we desire a method to produce \textit{optimized} draft graph structures.

To achieve this, we present a novel calibration algorithm to procedurally determine draft graph structures. A draft graph is defined by a set of $D$ \textbf{draft formulas}, where each formula is a set $\{(i, j)\}$, representing the tokens $\{c_{ij}\}$ that will populate the draft block. Intuitively, we wish to create drafts with assigned choices of $(i,j)$ so that we may select their unmasked token contents once the current timestep's probability distribution is available. Algorithm \ref{alg:calibration} details the calibration procedure, which involves running the vanilla dLLM on a small dataset and using the frequency of various sequences of $(i, j)$ to create optimized directed draft graphs. We also present a simplified pseudocode of Algorithm \ref{alg:calibration} in Appendix \ref{appendix:calibration-pseudocode} Algorithm \ref{alg:calibration-pseudo}.

In this work, we run calibration only once per model on a combined set of 25 samples each from MATH500 and MBPP. An example of a calibrated draft graph with $D=5$ is seen in Fig. \ref{fig:sample_graph_viz}. We provide further intuition for this calibration strategy in Appendix \ref{appendix:calibration-intuition} and discuss the costs of running this algorithm in Appendix \ref{appendix:calibration-time}, observing that it requires $\sim 10$ minutes to perform offline on a single GPU. In Appendix \ref{appendix:calibration-domain} we study the effect of changing the calibration dataset to a general text dataset, ShareGPT, observing comparable performance to the default calibration.

\definecolor{partHL}{RGB}{238,243,250}

\newcommand{\PartHL}[1]{%
  \Statex \colorbox{partHL}{\parbox{\dimexpr\linewidth-2\fboxsep\relax}{\textbf{#1}}}%
}

\begin{algorithm}[!tb]
\caption{Draft Graph Structure Determination via Offline Calibration}
\label{alg:calibration}
\begin{algorithmic}
\Statex \textbf{Inputs:} dLLM $\phi$, sample $Dataset$, `look-ahead' length for calibration $\mathcal{L}$, number of drafts budget $D$ 
\PartHL{\# Part 1: Collect calibration data}
\Statex \text{results} $\gets [\;]$
\For{sample $\in Dataset$}
    \State $X\gets \phi(sample)$ with:
    \State \hspace{\algorithmicindent} denoising interval $[T, 0]$
    \State \hspace{\algorithmicindent} recording of $p_{\text{TD}}(t)\quad \forall t$
    \Statex \hspace{1.25em}  \textbf{\# Rewind the generation of $X$}
    \Statex \hspace{1.25em} $\text{sequence} \gets [\;]$
    \For{every timestep $t \in [T, 0]$}
        \For{$t' \in [t, t+\mathcal{L}]$}
        \Statex \hspace{2.7em} \textbf{\# Unmasked tokens at current lookahead}
            \State $x_1, \ldots, x_k \gets$ tokens unmasked at $t'$
            \State Get $(i, j)$ for each token using (\ref{eq:i-select}, \ref{eq:j-select})
            \State Append $[(L = t' - t, \{(i, j)\})]$ to sequence
        \EndFor
    \EndFor
\EndFor
\Statex \textbf{\#} results = $\left[\,\text{seq}=\left[(L=k,\{(i,j)\})\right]\,\right]$
\smallskip 
\smallskip 
\PartHL{\# Part 2: All-possibilities-graph construction}
\For{$k \in 1, \ldots, \mathcal{L}$}
    \State Possibilities $\gets$ sequences in $results$ containing
    \State \hspace{1.25em} $L=k$ 
    \State For each $seq\in$ Possibilities, 
    \Statex \hspace{2.7em} \textbf{\# Each node is an unordered set $\{(i, j)\}$}
    \Statex \hspace{2.7em} count(node) := frequency of occurrence of node
    \Statex \hspace{2.7em} current\_node = $\bigcup_{e\in seq \; \mid \;e.L\leq k }\{(i, j)\}$ 
    \Statex \hspace{2.7em} previous\_node = $\bigcup_{e\in seq \; \mid \; e.L < k }\{(i, j)\}$ 
    \State top\_nodes = \{top-3 most frequent current\_nodes\}
    \State top\_in\_edges = \{top-3 most frequent previous\_nodes\}
    \State $G \gets$ top\_nodes, top\_edges. 
    \Statex \hspace{1.25em}  \textbf{\# Add additional edges}
    \State $G\gets \{(p, q) \; \forall \; p, q \in G \mid p\subset q\}$
\EndFor
\Statex \# $G \gets \text{nodes}=\{q_{kv}\}\text{ for } { v\in 1,2,3, \; k\in 1, \cdots, \mathcal{L}} $
\Statex \# $G \gets \text{edges}=\{(p_{k_1v}, q_{k_2u})\}$
\Statex \textbf{\# Each $q_{kv} = \{(i, j)\}$ is a draft formula}
\smallskip 
\smallskip
\PartHL{\# Part 3: Optimal subgraph selection}
\Statex $G' \gets$ all connected subgraphs of $G$ with size $D$
\Statex \hspace{2.4em} where $k=1, 2$ are root levels.
\Statex \textbf{\# Find the optimal subgraph $G^*$}
\Statex $G^* = \arg\max_{Q\in G'}\sum_{q\in Q}\left(count(q)\right)$
\Statex \textbf{$G^*=\{q_{kv}\}$, a set of $D$ draft formulas}
\smallskip
\Statex \textbf{Output} $G^*$
\end{algorithmic}
\end{algorithm}

\subsubsection{Draft Graph Pruning}
\label{sec:draft-pruning}
Draft graph calibration through Alg. \ref{alg:calibration} produces a structure with $D$ draft nodes. $D$ represents the number of draft blocks that will be appended to the sequence during verification. Although the inference of LLMs is memory-bound in general, it is beneficial to \textit{prune} the draft graph dynamically during inference to a smaller budget $D^*< D$ to save on computation. Since the draft nodes each represent sequences of tokens with a parent-child structure (Sec. \ref{sec:draft-graphs}), it is desirable to have a pruning metric that captures the joint probability of that sequence and the relationships between nodes. To this end, we define a metric based on the geometric mean of the token probabilities corresponding to each node. Suppose a node $q_i$ consists of tokens $c_{i1}, \cdots, c_{ik}$ with corresponding probabilities $p_{i1}, \cdots, p_{ik}$. We define the local score of a node via its geometric mean as,
\begin{align}
    localScore(q_i) &= GM(p_{i1}, \cdots, p_{ik})= \left(\prod_{j=1}^kp_{ij}\right)^{1/k}
\end{align}
and the aggregated score of its children as,
\begin{align}
    & childScore(q_i) = \nonumber  \\
    & GM\left(\left\{localScore(x)\; \text{for} \; x \in children(q_i)\right\}\right)
\end{align}
to compute the total score of the node,
\begin{align}
    score(q_i) = GM\left(localScore(q_i), childScore(q_i)\right)
\end{align}

$GM$ is used consistently as the aggregation metric to respect the multiplicative nature of probabilities. With this scoring metric, the top-$D^*$ nodes according to the current timestep's probability distribution can be flexibly chosen to be draft states. We see in Sec. \ref{sec:num-draft-blocks} that this metric enables near-complete recovery of the full draft graph's acceptance rate while reducing its computational overhead.

Thus, Spiffy's method involves first running the calibration algorithm offline to determine an optimized draft graph structure. Speculation then begins, using this graph configuration to determine the token candidates $c_{ij}$ that will appear in each draft block. The graph is pruned according to a desired budget and is then verified in parallel to accept draft states, resulting in accelerated inference. We provide a detailed profile of the overheads of Spiffy's drafting, pruning, and verification steps in Appendix \ref{appendix:overheads} and see that they are negligible compared to the model inference time.

\section{Experiments}
\label{sec:experiments}
Following the settings of \cite{fastdllmtrainingfreeaccelerationdiffusion} and \cite{wang2025revolutionizingtrado}, we demonstrate the acceleration provided by Spiffy for LLaDA-8B-Instruct \cite{llada}, Dream-7B-Instruct \cite{dream2025}, and SDAR-8B-Chat-b32 \cite{cheng2025sdarsynergisticdiffusionautoregressionparadigm} on standard coding and math tasks, GSM8K \cite{cobbe2021gsm8k}, MATH500 \cite{hendrycks2021measuringmathematicalproblemsolving}, MBPP \cite{mbppaustin2021program}, and HumanEval \cite{humanevalchen2021codex} with block size $32$. For LLaDA and Dream, the generation length is set to $256$ and the temperature to $0.0$ by default. We set SDAR's sampling temperature to $1.0$, following its recommended setting, and its generation length to $512$. We additionally compute the accuracy of the model outputs to validate Spiffy's lossless property. Prefix KV caching \cite{fastdllmtrainingfreeaccelerationdiffusion, ma2025dkv} is enabled in all settings and the baseline is set to be static unmasking with one token decoded per iteration. We compare this baseline to the same setting with dynamic unmasking with threshold $=0.9$ \cite{fastdllmtrainingfreeaccelerationdiffusion}. Spiffy is then enabled in this setting to demonstrate its combined speedup. 

To calibrate Spiffy, we use 25 samples from each of MBPP and MATH500, isolated from the test set to avoid contamination, and produce a single directed draft graph with $D=10$ for each model. This graph is then dynamically pruned to $D^*=3$ during inference. We calculate speedup with respect to the baseline in terms of increase in wall-clock token rate (TPS) and in terms of reduction in the Number of Function Evaluations (NFEs). 

\subsection{Main Experiment}
\label{sec:main-exp}
Table \ref{tab:main-exp}  displays the acceleration provided by Spiffy for LLaDA-8B-Instruct, Dream-7B-Instruct, and SDAR-8B-Chat-b32, and we see that across tasks and models, Spiffy enables NFE reductions of up to $9\times$ and token rate speedups of up to $6\times$ while preserving the model's accuracy. We include additional experiments on open-ended text-generation tasks from MMLU in Appendix \ref{appendix:open-text-exp} Table \ref{tab:open-text-exp}. 

\definecolor{nfeBlue}{HTML}{1F4E79}    
\definecolor{tpsOrange}{HTML}{C65D00}  
\definecolor{accTeal}{HTML}{0F766E}    
\newcommand{\cg}[1]{\textcolor{nfeBlue}{#1}}   
\newcommand{\co}[1]{\textcolor{tpsOrange}{#1}} 
\newcommand{\cb}[1]{\textcolor{accTeal}{#1}}   

\begin{table*}[!htb]
    \caption{Spiffy successfully results in \cg{NFE reductions} of up to $8.6\times$, with \co{TPS improvements} of up to $6.3\times$, while preserving \cb{accuracy} across dLLMs and tasks.}
    \centering
    \small
    \setlength{\tabcolsep}{5pt}
    \begin{tabular}{lcccc}
    \toprule
           & GSM8K & HumanEval & MATH500 & MBPP \\
    \midrule
    Baseline (LLaDA-8B-Instruct) & $\cg{1.00\times}\ (\co{1.00\times}),\ \cb{0.79}$ & $\cg{1.00\times}\ (\co{1.00\times}),\ \cb{0.41}$ & $\cg{1.00\times}\ (\co{1.00\times}),\ \cb{0.34}$ & $\cg{1.00\times}\ (\co{1.00\times}),\ \cb{0.36}$ \\
    + Dynamic Unmasking & $\cg{3.31\times}\ (\co{3.20\times}),\ \cb{0.79}$ & $\cg{3.51\times}\ (\co{2.89\times}),\ \cb{0.41}$ & $\cg{2.52\times}\ (\co{2.63\times}),\ \cb{0.35}$ & $\cg{5.73\times}\ (\co{5.01\times}),\ \cb{0.36}$ \\
    + Spiffy & $\cg{4.97\times}\ (\co{3.55\times}),\ \cb{0.79}$ & $\cg{5.80\times}\ (\co{3.22\times}),\ \cb{0.38}$ & $\cg{4.11\times}\ (\co{3.00\times}),\ \cb{0.35}$ & $\cg{8.58\times}\ (\co{5.23\times}),\ \cb{0.36}$ \\
    \midrule
    Baseline (Dream-7B-Instruct) & $\cg{1.00\times}\ (\co{1.00\times}),\ \cb{0.79}$ & $\cg{1.00\times}\ (\co{1.00\times}),\ \cb{0.54}$ & $\cg{1.00\times}\ (\co{1.00\times}),\ \cb{0.41}$ & $\cg{1.00\times}\ (\co{1.00\times}),\ \cb{0.49}$ \\
    + Dynamic Unmasking & $\cg{2.35\times}\ (\co{2.28\times}),\ \cb{0.78}$ & $\cg{2.68\times}\ (\co{2.73\times}),\ \cb{0.52}$ & $\cg{2.17\times}\ (\co{2.16\times}),\ \cb{0.42}$ & $\cg{2.45\times}\ (\co{2.63\times}),\ \cb{0.50}$ \\
    + Spiffy & $\cg{3.76\times}\ (\co{2.56\times}),\ \cb{0.79}$ & $\cg{4.42\times}\ (\co{2.90\times}),\ \cb{0.55}$ & $\cg{3.57\times}\ (\co{2.57\times}),\ \cb{0.42}$ & $\cg{4.00\times}\ (\co{2.97\times}),\ \cb{0.49}$ \\
    \midrule 
    Baseline (SDAR-8B-Chat-b32) & $\cg{1.00\times}\ (\co{1.00\times}),\ \cb{0.90}$ & $\cg{1.00\times}\ (\co{1.00\times}),\ \cb{0.68}$ & $\cg{1.00\times}\ (\co{1.00\times}),\ \cb{0.49}$ & $\cg{1.00\times}\ (\co{1.00\times}),\ \cb{0.26}$ \\
    + Dynamic Unmasking & $\cg{6.71\times}\ (\co{5.67\times}),\ \cb{0.88}$ & $\cg{5.54\times}\ (\co{4.24\times}),\ \cb{0.71}$ & $\cg{4.38\times}\ (\co{4.11\times}),\ \cb{0.47}$ & $\cg{3.99\times}\ (\co{3.45\times}),\ \cb{0.24}$ \\
    + Spiffy & $\cg{8.25\times}\ (\co{6.28\times}),\ \cb{0.88}$ & $\cg{6.96\times}\ (\co{4.77\times}),\ \cb{0.71}$ & $\cg{5.46\times}\ (\co{4.60\times}),\ \cb{0.47}$ & $\cg{5.29\times}\ (\co{4.21\times}),\ \cb{0.25}$ \\
    \bottomrule
    
    \end{tabular}
    
\label{tab:main-exp}
\end{table*}

\subsection{Number of Draft Blocks}
\label{sec:num-draft-blocks}
\begin{figure}[ht]
    \centering
    \includegraphics[width=0.9\columnwidth]{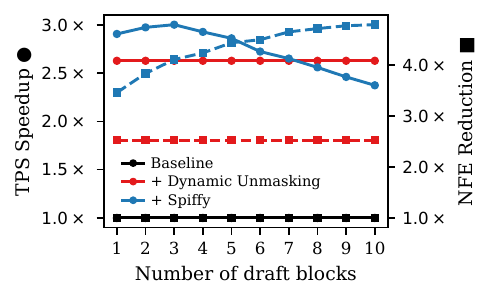}
    \caption{Spiffy's draft pruning procedure achieves a balance between reduction in NFEs (squares) and wall-clock token rate speedup (circles) depending on the preferred metric. The baseline is LLaDA-8B-Instruct on MATH500 with prefix-caching.}
    \label{fig:num_drafts}
\end{figure}

Using the draft pruning procedure in Sec. \ref{sec:draft-pruning}, we demonstrate in Fig. \ref{fig:num_drafts} the effect of the number of draft blocks on TPS speedup and NFE reduction. Using a higher number of draft blocks will result in more acceptances and further reductions in NFEs, while fewer draft blocks will result in decreased acceptance, but lower computational costs. By varying the draft budget using our draft pruning procedure, we can strike a balance between these metrics according to the system's requirements.

\subsection{Effect of Temperature}
\label{sec:exp-temperature}
\begin{figure}[ht]
    \centering
    \includegraphics[width=0.9\columnwidth]{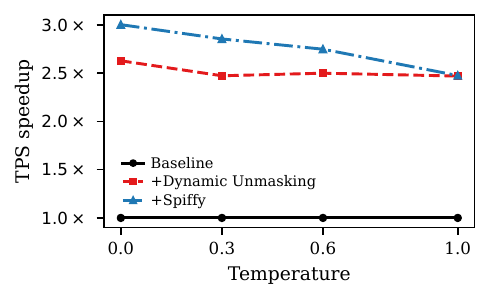}
    \caption{Spiffy remains effective at speeding up LLaDA-8B-Instruct inference even with increasing sampling temperature.}
    \label{fig:temp}
\end{figure}

In Fig. \ref{fig:temp}, we study the effect of temperature on Spiffy's speedup for LLaDA-8B-Instruct with prefix caching. We reuse the calibrated draft graph generated with temperature $=0.0$ for all settings and see that with increasing sampling temperature, draft acceptance is more challenging, yet Spiffy can achieve higher TPS than the baseline. We expect that re-calibrating the draft graph for the desired temperature setting would improve performance further.

\section{Discussion and Conclusion}
Spiffy is a novel speculative decoding algorithm for diffusion LLMs that uses structured draft graphs to capture token unmasking dynamics. In this work, we define directed draft graphs that take advantage of the bidirectional nature of dLLM decoding to maximize acceptance. Verification of these draft graphs occurs in parallel and results in preserved model accuracy. We develop a novel, inexpensive calibration algorithm to optimize the draft graph structures and a pruning procedure that reduces its computational requirements. This framework is implemented via auto-speculation, eliminating the overheads of training and deploying an independent draft model. Spiffy is shown to be effective for multiple open-source dLLM model families, LLaDA, Dream, and SDAR, evaluated on multiple tasks. We further ablate on the impact of the number of draft blocks and temperature on Spiffy's acceleration benefits. Future investigations could explore alternate calibration and pruning metrics, and substituting Spiffy's auto-speculation implementation with the use of an auxiliary draft model. We demonstrate that Spiffy's formulation generalizes to arbitrary denoising rates and as a result, as future dLLMs require decreasing numbers of denoising steps, Spiffy will remain an effective acceleration solution.

\section*{Impact Statement}

This paper presents work whose goal is to advance the field of Machine Learning. There are many potential societal consequences of our work, none which we feel must be specifically highlighted here.

\newpage

\bibliography{main}

\begin{thebibliography}{58}
\providecommand{\natexlab}[1]{#1}
\providecommand{\url}[1]{\texttt{#1}}
\expandafter\ifx\csname urlstyle\endcsname\relax
  \providecommand{\doi}[1]{doi: #1}\else
  \providecommand{\doi}{doi: \begingroup \urlstyle{rm}\Url}\fi

\bibitem[Arriola et~al.(2025)Arriola, Gokaslan, Chiu, Yang, Qi, Han, Sahoo, and Kuleshov]{arriola2025block}
Arriola, M., Gokaslan, A., Chiu, J.~T., Yang, Z., Qi, Z., Han, J., Sahoo, S.~S., and Kuleshov, V.
\newblock Block diffusion: Interpolating between autoregressive and diffusion language models.
\newblock \emph{arXiv preprint arXiv:2503.09573}, 2025.

\bibitem[Austin et~al.(2021{\natexlab{a}})Austin, Johnson, Ho, Tarlow, and Van Den~Berg]{austin2021structured}
Austin, J., Johnson, D.~D., Ho, J., Tarlow, D., and Van Den~Berg, R.
\newblock Structured denoising diffusion models in discrete state-spaces.
\newblock \emph{Advances in neural information processing systems}, 34:\penalty0 17981--17993, 2021{\natexlab{a}}.

\bibitem[Austin et~al.(2021{\natexlab{b}})Austin, Odena, Nye, Bosma, Michalewski, Dohan, Jiang, Cai, Terry, Le, et~al.]{mbppaustin2021program}
Austin, J., Odena, A., Nye, M., Bosma, M., Michalewski, H., Dohan, D., Jiang, E., Cai, C., Terry, M., Le, Q., et~al.
\newblock Program synthesis with large language models.
\newblock \emph{arXiv preprint arXiv:2108.07732}, 2021{\natexlab{b}}.

\bibitem[Bie et~al.(2025)Bie, Cao, Chen, Du, Gong, Gong, Gu, Hu, Huang, Lan, et~al.]{bie2025llada2}
Bie, T., Cao, M., Chen, K., Du, L., Gong, M., Gong, Z., Gu, Y., Hu, J., Huang, Z., Lan, Z., et~al.
\newblock Llada2. 0: Scaling up diffusion language models to 100b.
\newblock \emph{arXiv preprint arXiv:2512.15745}, 2025.

\bibitem[{Black Forest Labs} et~al.(2025){Black Forest Labs}, Batifol, Blattmann, Boesel, Consul, Diagne, Dockhorn, English, English, Esser, et~al.]{labs2025flux}
{Black Forest Labs}, Batifol, S., Blattmann, A., Boesel, F., Consul, S., Diagne, C., Dockhorn, T., English, J., English, Z., Esser, P., et~al.
\newblock Flux. 1 kontext: Flow matching for in-context image generation and editing in latent space.
\newblock \emph{arXiv preprint arXiv:2506.15742}, 2025.

\bibitem[Brown et~al.(2020)Brown, Mann, Ryder, Subbiah, Kaplan, Dhariwal, Neelakantan, Shyam, Sastry, Askell, et~al.]{brown2020language}
Brown, T., Mann, B., Ryder, N., Subbiah, M., Kaplan, J.~D., Dhariwal, P., Neelakantan, A., Shyam, P., Sastry, G., Askell, A., et~al.
\newblock Language models are few-shot learners.
\newblock \emph{Advances in neural information processing systems}, 33:\penalty0 1877--1901, 2020.

\bibitem[Cai et~al.(2024)Cai, Li, Geng, Peng, Lee, Chen, and Dao]{cai2024medusa}
Cai, T., Li, Y., Geng, Z., Peng, H., Lee, J.~D., Chen, D., and Dao, T.
\newblock Medusa: Simple llm inference acceleration framework with multiple decoding heads.
\newblock \emph{arXiv preprint arXiv:2401.10774}, 2024.

\bibitem[Chen et~al.(2023)Chen, Borgeaud, Irving, Lespiau, Sifre, and Jumper]{chen2023acceleratinglargelanguagemodel}
Chen, C., Borgeaud, S., Irving, G., Lespiau, J.-B., Sifre, L., and Jumper, J.
\newblock Accelerating large language model decoding with speculative sampling, 2023.
\newblock URL \url{https://arxiv.org/abs/2302.01318}.

\bibitem[Chen et~al.(2021)Chen, Tworek, Jun, Yuan, Pinto, Kaplan, Edwards, Burda, Joseph, Brockman, et~al.]{humanevalchen2021codex}
Chen, M., Tworek, J., Jun, H., Yuan, Q., Pinto, H. P. D.~O., Kaplan, J., Edwards, H., Burda, Y., Joseph, N., Brockman, G., et~al.
\newblock Evaluating large language models trained on code.
\newblock \emph{arXiv preprint arXiv:2107.03374}, 2021.

\bibitem[Cheng et~al.(2025)Cheng, Bian, Liu, Zhang, Yao, Tian, Wang, Guo, Chen, Qi, and Zhou]{cheng2025sdarsynergisticdiffusionautoregressionparadigm}
Cheng, S., Bian, Y., Liu, D., Zhang, L., Yao, Q., Tian, Z., Wang, W., Guo, Q., Chen, K., Qi, B., and Zhou, B.
\newblock Sdar: A synergistic diffusion-autoregression paradigm for scalable sequence generation, 2025.
\newblock URL \url{https://arxiv.org/abs/2510.06303}.

\bibitem[Christopher et~al.(2024)Christopher, Bartoldson, Ben-Nun, Cardei, Kailkhura, and Fioretto]{christopher2024speculative}
Christopher, J.~K., Bartoldson, B.~R., Ben-Nun, T., Cardei, M., Kailkhura, B., and Fioretto, F.
\newblock Speculative diffusion decoding: Accelerating language generation through diffusion.
\newblock \emph{arXiv preprint arXiv:2408.05636}, 2024.

\bibitem[Cobbe et~al.(2021)Cobbe, Kosaraju, Bavarian, Chen, Jun, Kaiser, Plappert, Tworek, Hilton, Nakano, Hesse, and Schulman]{cobbe2021gsm8k}
Cobbe, K., Kosaraju, V., Bavarian, M., Chen, M., Jun, H., Kaiser, L., Plappert, M., Tworek, J., Hilton, J., Nakano, R., Hesse, C., and Schulman, J.
\newblock Training verifiers to solve math word problems.
\newblock \emph{arXiv preprint arXiv:2110.14168}, 2021.

\bibitem[De~Bortoli et~al.(2025)De~Bortoli, Galashov, Gretton, and Doucet]{de2025accelerated}
De~Bortoli, V., Galashov, A., Gretton, A., and Doucet, A.
\newblock Accelerated diffusion models via speculative sampling.
\newblock \emph{arXiv preprint arXiv:2501.05370}, 2025.

\bibitem[Deschenaux \& Gulcehre(2025)Deschenaux and Gulcehre]{deschenaux2025beyond}
Deschenaux, J. and Gulcehre, C.
\newblock Beyond autoregression: Fast {LLM}s via self-distillation through time.
\newblock In \emph{The Thirteenth International Conference on Learning Representations}, 2025.
\newblock URL \url{https://openreview.net/forum?id=uZ5K4HeNwd}.

\bibitem[Devlin et~al.(2019)Devlin, Chang, Lee, and Toutanova]{bert}
Devlin, J., Chang, M.-W., Lee, K., and Toutanova, K.
\newblock Bert: Pre-training of deep bidirectional transformers for language understanding.
\newblock In \emph{Proceedings of the 2019 conference of the North American chapter of the association for computational linguistics: human language technologies, volume 1 (long and short papers)}, pp.\  4171--4186, 2019.

\bibitem[Du et~al.(2020)Du, Li, and Mordatch]{du2020compositional}
Du, Y., Li, S., and Mordatch, I.
\newblock Compositional visual generation with energy based models.
\newblock \emph{Advances in Neural Information Processing Systems}, 33:\penalty0 6637--6647, 2020.

\bibitem[Esser et~al.(2024)Esser, Kulal, Blattmann, Entezari, M{\"u}ller, Saini, Levi, Lorenz, Sauer, Boesel, Podell, Dockhorn, English, and Rombach]{esser2024scaling}
Esser, P., Kulal, S., Blattmann, A., Entezari, R., M{\"u}ller, J., Saini, H., Levi, Y., Lorenz, D., Sauer, A., Boesel, F., Podell, D., Dockhorn, T., English, Z., and Rombach, R.
\newblock Scaling rectified flow transformers for high-resolution image synthesis.
\newblock In \emph{Forty-first International Conference on Machine Learning}, 2024.
\newblock URL \url{https://openreview.net/forum?id=FPnUhsQJ5B}.

\bibitem[Fu et~al.(2024{\natexlab{a}})Fu, Bailis, Stoica, and Zhang]{fu2024break}
Fu, Y., Bailis, P., Stoica, I., and Zhang, H.
\newblock Break the sequential dependency of llm inference using lookahead decoding.
\newblock \emph{arXiv preprint arXiv:2402.02057}, 2024{\natexlab{a}}.

\bibitem[Fu et~al.(2024{\natexlab{b}})Fu, Bailis, Stoica, and Zhang]{lookaheadfu2024break}
Fu, Y., Bailis, P., Stoica, I., and Zhang, H.
\newblock Break the sequential dependency of llm inference using lookahead decoding.
\newblock \emph{arXiv preprint arXiv:2402.02057}, 2024{\natexlab{b}}.

\bibitem[Gao et~al.(2025)Gao, Ji, Wang, Qi, Xu, and Zhang]{gao2025selfspeculativedecodingdiffusion}
Gao, Y., Ji, Z., Wang, Y., Qi, B., Xu, H., and Zhang, L.
\newblock Self speculative decoding for diffusion large language models, 2025.
\newblock URL \url{https://arxiv.org/abs/2510.04147}.

\bibitem[Goel et~al.(2024)Goel, Gagrani, Jeon, Park, Lee, and Lott]{goel2024direct}
Goel, R., Gagrani, M., Jeon, W., Park, J., Lee, M., and Lott, C.
\newblock Direct alignment of draft model for speculative decoding with chat-fine-tuned llms.
\newblock \emph{arXiv preprint arXiv:2403.00858}, 2024.

\bibitem[Gong et~al.(2025)Gong, Zhang, Zheng, Gu, Jaitly, Kong, and Zhang]{gong2025diffucoder}
Gong, S., Zhang, R., Zheng, H., Gu, J., Jaitly, N., Kong, L., and Zhang, Y.
\newblock Diffucoder: Understanding and improving masked diffusion models for code generation.
\newblock \emph{arXiv preprint arXiv:2506.20639}, 2025.

\bibitem[Google(2025)]{Google_2025}
Google.
\newblock Gemini diffusion, 2025.
\newblock Accessed: 2025-07-25.

\bibitem[Hendrycks et~al.(2021)Hendrycks, Burns, Kadavath, Arora, Basart, Tang, Song, and Steinhardt]{hendrycks2021measuringmathematicalproblemsolving}
Hendrycks, D., Burns, C., Kadavath, S., Arora, A., Basart, S., Tang, E., Song, D., and Steinhardt, J.
\newblock Measuring mathematical problem solving with the math dataset, 2021.
\newblock URL \url{https://arxiv.org/abs/2103.03874}.

\bibitem[Ho et~al.(2020)Ho, Jain, and Abbeel]{ho2020denoising}
Ho, J., Jain, A., and Abbeel, P.
\newblock Denoising diffusion probabilistic models.
\newblock \emph{Advances in neural information processing systems}, 33:\penalty0 6840--6851, 2020.

\bibitem[Hong et~al.(2025)Hong, Yu, Ye, Huang, Zheng, Zhang, Wang, and Yao]{hong2025wide}
Hong, F., Yu, G., Ye, Y., Huang, H., Zheng, H., Zhang, Y., Wang, Y., and Yao, J.
\newblock Wide-in, narrow-out: Revokable decoding for efficient and effective dllms.
\newblock \emph{arXiv preprint arXiv:2507.18578}, 2025.

\bibitem[Hu et~al.(2025)Hu, Das, Sadigh, and Anari]{hu2025diffusion}
Hu, H., Das, A., Sadigh, D., and Anari, N.
\newblock Diffusion models are secretly exchangeable: Parallelizing ddpms via autospeculation.
\newblock \emph{arXiv preprint arXiv:2505.03983}, 2025.

\bibitem[{Inception Labs} et~al.(2025){Inception Labs}, Khanna, Kharbanda, Li, Varma, Wang, Birnbaum, Luo, Miraoui, Palrecha, Ermon, Grover, and Kuleshov]{mercury}
{Inception Labs}, Khanna, S., Kharbanda, S., Li, S., Varma, H., Wang, E., Birnbaum, S., Luo, Z., Miraoui, Y., Palrecha, A., Ermon, S., Grover, A., and Kuleshov, V.
\newblock Mercury: Ultra-fast language models based on diffusion, 2025.
\newblock URL \url{https://arxiv.org/abs/2506.17298}.

\bibitem[Israel et~al.(2025)Israel, Broeck, and Grover]{israel2025acceleratingdiffusionllmsadaptive}
Israel, D., Broeck, G. V.~d., and Grover, A.
\newblock Accelerating diffusion llms via adaptive parallel decoding.
\newblock \emph{arXiv preprint arXiv:2506.00413}, 2025.

\bibitem[Jeon et~al.(2024)Jeon, Gagrani, Goel, Park, Lee, and Lott]{jeon2024recursive}
Jeon, W., Gagrani, M., Goel, R., Park, J., Lee, M., and Lott, C.
\newblock Recursive speculative decoding: Accelerating llm inference via sampling without replacement.
\newblock \emph{arXiv preprint arXiv:2402.14160}, 2024.

\bibitem[Leviathan et~al.(2023)Leviathan, Kalman, and Matias]{leviathan2023fast}
Leviathan, Y., Kalman, M., and Matias, Y.
\newblock Fast inference from transformers via speculative decoding.
\newblock In \emph{International Conference on Machine Learning}, pp.\  19274--19286. PMLR, 2023.

\bibitem[Li et~al.(2025{\natexlab{a}})Li, Kallidromitis, Bansal, Gokul, Kato, Kozuka, Kuen, Lin, Chang, and Grover]{li2025lavida}
Li, S., Kallidromitis, K., Bansal, H., Gokul, A., Kato, Y., Kozuka, K., Kuen, J., Lin, Z., Chang, K.-W., and Grover, A.
\newblock Lavida: A large diffusion language model for multimodal understanding.
\newblock \emph{arXiv preprint arXiv:2505.16839}, 2025{\natexlab{a}}.

\bibitem[Li et~al.(2024)Li, Wei, Zhang, and Zhang]{li2024eagle}
Li, Y., Wei, F., Zhang, C., and Zhang, H.
\newblock Eagle-2: Faster inference of language models with dynamic draft trees.
\newblock \emph{arXiv preprint arXiv:2406.16858}, 2024.

\bibitem[Li et~al.(2025{\natexlab{b}})Li, Wei, Zhang, and Zhang]{li2025eagle3scalinginferenceacceleration}
Li, Y., Wei, F., Zhang, C., and Zhang, H.
\newblock Eagle-3: Scaling up inference acceleration of large language models via training-time test, 2025{\natexlab{b}}.
\newblock URL \url{https://arxiv.org/abs/2503.01840}.

\bibitem[Lin et~al.(2025)Lin, Yi, Yang, Li, Yu, Lu, and Xiao]{lin2025bita}
Lin, F., Yi, H., Yang, Y., Li, H., Yu, X., Lu, G., and Xiao, R.
\newblock Bita: Bi-directional tuning for lossless acceleration in large language models.
\newblock \emph{Expert Systems with Applications}, 279:\penalty0 127305, 2025.

\bibitem[Liu et~al.(2025{\natexlab{a}})Liu, Yang, Zhang, Chen, Zou, Wei, Wang, and Zhang]{liu2025dllm}
Liu, Z., Yang, Y., Zhang, Y., Chen, J., Zou, C., Wei, Q., Wang, S., and Zhang, L.
\newblock dllm-cache: Accelerating diffusion large language models with adaptive caching.
\newblock \emph{arXiv preprint arXiv:2506.06295}, 2025{\natexlab{a}}.

\bibitem[Liu et~al.(2025{\natexlab{b}})Liu, Yang, Zhang, Chen, Zou, Wei, Wang, and Zhang]{liu2025dllmcacheacceleratingdiffusionlarge}
Liu, Z., Yang, Y., Zhang, Y., Chen, J., Zou, C., Wei, Q., Wang, S., and Zhang, L.
\newblock dllm-cache: Accelerating diffusion large language models with adaptive caching, 2025{\natexlab{b}}.
\newblock URL \url{https://arxiv.org/abs/2506.06295}.

\bibitem[Lou et~al.(2023)Lou, Meng, and Ermon]{lou2023discrete}
Lou, A., Meng, C., and Ermon, S.
\newblock Discrete diffusion modeling by estimating the ratios of the data distribution.
\newblock \emph{arXiv preprint arXiv:2310.16834}, 2023.

\bibitem[Luxembourg et~al.(2025)Luxembourg, Permuter, and Nachmani]{luxembourg2025plan}
Luxembourg, O., Permuter, H., and Nachmani, E.
\newblock Plan for speed--dilated scheduling for masked diffusion language models.
\newblock \emph{arXiv preprint arXiv:2506.19037}, 2025.

\bibitem[Ma et~al.(2025)Ma, Yu, Fang, and Wang]{ma2025dkv}
Ma, X., Yu, R., Fang, G., and Wang, X.
\newblock dkv-cache: The cache for diffusion language models.
\newblock \emph{arXiv preprint arXiv:2505.15781}, 2025.

\bibitem[Miao et~al.(2024)Miao, Oliaro, Zhang, Cheng, Wang, Zhang, Wong, Zhu, Yang, Shi, et~al.]{miao2024specinfer}
Miao, X., Oliaro, G., Zhang, Z., Cheng, X., Wang, Z., Zhang, Z., Wong, R. Y.~Y., Zhu, A., Yang, L., Shi, X., et~al.
\newblock Specinfer: Accelerating large language model serving with tree-based speculative inference and verification.
\newblock In \emph{Proceedings of the 29th ACM International Conference on Architectural Support for Programming Languages and Operating Systems, Volume 3}, pp.\  932--949, 2024.

\bibitem[Ni \& team(2025)Ni and team]{ni2025openmoe2}
Ni, J. and team.
\newblock Openmoe 2: Sparse diffusion language models.
\newblock \url{https://github.com/JinjieNi/OpenMoE2}, 2025.

\bibitem[Nie et~al.(2025)Nie, Zhu, You, Zhang, Ou, Hu, Zhou, Lin, Wen, and Li]{llada}
Nie, S., Zhu, F., You, Z., Zhang, X., Ou, J., Hu, J., Zhou, J., Lin, Y., Wen, J.-R., and Li, C.
\newblock Large language diffusion models, 2025.
\newblock URL \url{https://arxiv.org/abs/2502.09992}.

\bibitem[Qian et~al.(2026)Qian, Su, Hu, Zhang, Deng, Zhao, and Zhang]{qian2026d3llm}
Qian, Y.-Y., Su, J., Hu, L., Zhang, P., Deng, Z., Zhao, P., and Zhang, H.
\newblock d3llm: Ultra-fast diffusion llm using pseudo-trajectory distillation.
\newblock \emph{arXiv preprint arXiv:2601.07568}, 2026.

\bibitem[Saharia et~al.(2022)Saharia, Chan, Saxena, Li, Whang, Denton, Ghasemipour, Gontijo-Lopes, Ayan, Salimans, Ho, Fleet, and Norouzi]{saharia2022photorealistic}
Saharia, C., Chan, W., Saxena, S., Li, L., Whang, J., Denton, E., Ghasemipour, S. K.~S., Gontijo-Lopes, R., Ayan, B.~K., Salimans, T., Ho, J., Fleet, D.~J., and Norouzi, M.
\newblock Photorealistic text-to-image diffusion models with deep language understanding.
\newblock In Oh, A.~H., Agarwal, A., Belgrave, D., and Cho, K. (eds.), \emph{Advances in Neural Information Processing Systems}, 2022.
\newblock URL \url{https://openreview.net/forum?id=08Yk-n5l2Al}.

\bibitem[Sahoo et~al.(2024)Sahoo, Arriola, Schiff, Gokaslan, Marroquin, Chiu, Rush, and Kuleshov]{sahoo2024simple}
Sahoo, S., Arriola, M., Schiff, Y., Gokaslan, A., Marroquin, E., Chiu, J., Rush, A., and Kuleshov, V.
\newblock Simple and effective masked diffusion language models.
\newblock \emph{Advances in Neural Information Processing Systems}, 37:\penalty0 130136--130184, 2024.

\bibitem[Shaul et~al.(2024)Shaul, Gat, Havasi, Severo, Sriram, Holderrieth, Karrer, Lipman, and Chen]{shaul2024flow}
Shaul, N., Gat, I., Havasi, M., Severo, D., Sriram, A., Holderrieth, P., Karrer, B., Lipman, Y., and Chen, R.~T.
\newblock Flow matching with general discrete paths: A kinetic-optimal perspective.
\newblock \emph{arXiv preprint arXiv:2412.03487}, 2024.

\bibitem[Shi et~al.(2024)Shi, Han, Wang, Doucet, and Titsias]{shi2024simplified}
Shi, J., Han, K., Wang, Z., Doucet, A., and Titsias, M.
\newblock Simplified and generalized masked diffusion for discrete data.
\newblock \emph{Advances in neural information processing systems}, 37:\penalty0 103131--103167, 2024.

\bibitem[Singhal et~al.(2025)Singhal, Horvitz, Teehan, Ren, Yu, McKeown, and Ranganath]{singhal2025general}
Singhal, R., Horvitz, Z., Teehan, R., Ren, M., Yu, Z., McKeown, K., and Ranganath, R.
\newblock A general framework for inference-time scaling and steering of diffusion models.
\newblock \emph{arXiv preprint arXiv:2501.06848}, 2025.

\bibitem[Sohl-Dickstein et~al.(2015)Sohl-Dickstein, Weiss, Maheswaranathan, and Ganguli]{pmlr-v37-sohl-dickstein15}
Sohl-Dickstein, J., Weiss, E., Maheswaranathan, N., and Ganguli, S.
\newblock Deep unsupervised learning using nonequilibrium thermodynamics.
\newblock In Bach, F. and Blei, D. (eds.), \emph{Proceedings of the 32nd International Conference on Machine Learning}, volume~37 of \emph{Proceedings of Machine Learning Research}, pp.\  2256--2265, Lille, France, 07--09 Jul 2015. PMLR.
\newblock URL \url{https://proceedings.mlr.press/v37/sohl-dickstein15.html}.

\bibitem[Song et~al.(2020)Song, Sohl-Dickstein, Kingma, Kumar, Ermon, and Poole]{song2020score}
Song, Y., Sohl-Dickstein, J., Kingma, D.~P., Kumar, A., Ermon, S., and Poole, B.
\newblock Score-based generative modeling through stochastic differential equations.
\newblock \emph{arXiv preprint arXiv:2011.13456}, 2020.

\bibitem[Song et~al.(2025)Song, Zhang, Luo, Gao, Xia, Luo, Li, Yang, Yu, Qu, Fu, Su, Zhang, Huang, Wang, Yan, Jia, Liu, Ma, Zhang, Wu, and Zhou]{song2025seeddiffusionlargescalediffusion}
Song, Y., Zhang, Z., Luo, C., Gao, P., Xia, F., Luo, H., Li, Z., Yang, Y., Yu, H., Qu, X., Fu, Y., Su, J., Zhang, G., Huang, W., Wang, M., Yan, L., Jia, X., Liu, J., Ma, W.-Y., Zhang, Y.-Q., Wu, Y., and Zhou, H.
\newblock Seed diffusion: A large-scale diffusion language model with high-speed inference, 2025.
\newblock URL \url{https://arxiv.org/abs/2508.02193}.

\bibitem[Wang et~al.(2025)Wang, Yang, Li, Tian, Shen, and Wang]{wang2025revolutionizingtrado}
Wang, Y., Yang, L., Li, B., Tian, Y., Shen, K., and Wang, M.
\newblock Revolutionizing reinforcement learning framework for diffusion large language models.
\newblock \emph{arXiv preprint arXiv:2509.06949}, 2025.

\bibitem[Wang et~al.(2024)Wang, Zhang, Ding, Yang, Li, and Xiang]{wang2024continuous}
Wang, Z., Zhang, R., Ding, K., Yang, Q., Li, F., and Xiang, S.
\newblock Continuous speculative decoding for autoregressive image generation.
\newblock \emph{arXiv preprint arXiv:2411.11925}, 2024.

\bibitem[Wu et~al.(2025)Wu, Zhang, Xue, Liu, Diao, Zhu, Luo, Han, and Xie]{fastdllmtrainingfreeaccelerationdiffusion}
Wu, C., Zhang, H., Xue, S., Liu, Z., Diao, S., Zhu, L., Luo, P., Han, S., and Xie, E.
\newblock Fast-dllm: Training-free acceleration of diffusion llm by enabling kv cache and parallel decoding, 2025.
\newblock URL \url{https://arxiv.org/abs/2505.22618}.

\bibitem[Xie et~al.(2025)Xie, Ye, Zheng, Gao, Dong, Wu, Zhao, Gong, Jiang, Li, et~al.]{xie2025dream}
Xie, Z., Ye, J., Zheng, L., Gao, J., Dong, J., Wu, Z., Zhao, X., Gong, S., Jiang, X., Li, Z., et~al.
\newblock Dream-coder 7b: An open diffusion language model for code.
\newblock \emph{arXiv preprint arXiv:2509.01142}, 2025.

\bibitem[Ye et~al.(2025)Ye, Xie, Zheng, Gao, Wu, Jiang, Li, and Kong]{dream2025}
Ye, J., Xie, Z., Zheng, L., Gao, J., Wu, Z., Jiang, X., Li, Z., and Kong, L.
\newblock Dream 7b, 2025.
\newblock URL \url{https://hkunlp.github.io/blog/2025/dream}.

\bibitem[Zhang et~al.(2023)Zhang, Wang, Li, Shou, Chen, Chen, and Mehrotra]{zhang2023draft}
Zhang, J., Wang, J., Li, H., Shou, L., Chen, K., Chen, G., and Mehrotra, S.
\newblock Draft \& verify: Lossless large language model acceleration via self-speculative decoding.
\newblock \emph{arXiv preprint arXiv:2309.08168}, 2023.

\end{thebibliography}
\bibliographystyle{icml_2026_arxiv}

\newpage
\appendix
\onecolumn
\section{Losslessness}
\label{appendix:proof-lossless}

Let the current state of the sequence be $X(t_0)=X';X_k(t_0)$. We wish to show that when the denoising progresses from $t_0\rightarrow t_0-2$, we achieve the same result whether we had used only the true (target) probabilities or accepted a draft block using Spiffy and skipped to timestep $t_0-2$. 

First, define the process of sampling the next state of a block $X$ from some probability distribution $p$ as a function \texttt{Next($X$, $p$)}. This is simply the process of using $p$ to performing ranking and picking of token positions to be unmasked in $X$ described in \texttt{Steps 1-4} of Alg. \ref{alg:verification} and is common to both speculative and non-speculative implementations. 

\subsection{Proof}
\label{appendix:proof-formal}

\textbf{Vanilla decoding output: } Let $X^*_k(t_0-2)$ be the final state of the sequence with no speculation. From $t_0\rightarrow t_0-1$, we may obtain the next state as: 
\begin{align}
    &p_{\text{TD}}(t_0-1) \gets \nonumber \\
    &\quad p_{\theta}\left(X_k(t_0-1) \mid X'; X_k(t_0)\right) \\
    &X_k(t_0-1) \gets \nonumber \\
    &\quad \texttt{Next}\left(X_k(t_0), p_{\text{TD}}(t_0-1) \right) 
    \label{eq:app-sample-t-1}
\end{align}
This state becomes the true state of the block, and inference is performed on it once more by the dLLM to advance to $t_0-2$ 
\begin{align}
    &p_{\text{TD}}(t_0-2) \gets \nonumber \\
    &\quad p_\theta \left(X_k(t_0-2) \mid X'; X_k(t_0-1)\right) \label{eq:p_tm} \\
    &X_k^*(t_0-2) \gets \nonumber \\
    &\quad \texttt{Next}\left(X_k(t_0-1), p_{\text{TD}}(t_0-2) \right) \label{eq:app-sample-t-2} 
\end{align}

\textbf{Spiffy output: } Let $X_k^+(t_0-2)$ be the state of the sequence if we first accept a draft block $\hat{X}^m_k$ at timestep $t_m$ using Alg. \ref{alg:verification}, then treat it as the true block, and then continue to the next iteration of the algorithm where we perform unmasking once more. 

First, in the case of speculation, we also sample $X_k(t_0-1)$ as usual in \texttt{Steps 1-4}, as is done in \eqref{eq:app-sample-t-1}. The next step of the algorithm is to check for acceptance or rejection. 

\textbf{Case 1 (rejection)}: For any of the drafts $m\in \{1, \cdots, D\}$, if $\hat{X}_k^m$ is rejected, the algorithm continues to the next available draft and checks again for acceptance or rejection. If all drafts are rejected, we simply break and output $X_k(t_0-1)$ as the true block state. At this point, in the next iteration, we will call the dLLM once more and this state $X_k(t_0-1)$ will be used to compute $p_{\text{TD}}$ as is done in \eqref{eq:p_tm}. Thus, we obtain $X_k^+(t_0-2)$ in the same way, and from the same distribution, as $X_k^*(t_0-2)$ and thus they are equal. 

\textbf{Case 2 (acceptance)}: If $\hat{X}_k^m$ is accepted, we now set $p_{\text{TD}} \gets p^m_{\text{DD}}(t_m-1)$ and restart the algorithm. In this next step, we sample from this distribution to obtain  
\begin{align}
    &X_k^+(t_0-2) \gets \nonumber \\
    &\quad \quad \texttt{Next}\left(\hat{X}_k^m(t_m-1), p_{\text{DD}}^m(t_m) \right) 
\end{align}

Since we previously had acceptance, we first have that $t_m==t_0-1$. Moreover, we have that $\hat{X}_k^m == X_k(t_0-1)$ from the acceptance condition. So we see that:  
$$p_{\text{DD}}^m(t_m-1) = p_{\text{DD}}^m(t_0-2)$$
and, since by definition 
$$p_{\text{DD}}^m(t_m-1) = p_\theta\left(X_k(t_m-1)\mid X';\hat{X}_k^m\right)$$
we have that 
\begin{align}
    p_{\text{DD}}^m(t_0-2) &= p_\theta\left(X_k(t_0-2)\mid X';X_k(t_0-1)\right)\\
    &=p_{\text{TD}}(t_0 -2) 
\end{align}
So, in reality, we sample 
\begin{align}
    &p_{\text{DD}}^m(t_0-2) \gets \nonumber \\
    &\quad p_\theta\left(X_k(t_0-2)\mid X';X_k(t_0-1)\right) \nonumber \\
    &\quad=p_{\text{TD}}(t_0 -2) \\
    &X_k^+(t_0-2) \gets \nonumber \\
    &\quad \quad \texttt{Next}\left(X_k^m(t_0-2), p_{\text{DD}}^m(t_0-2) \right) \nonumber \\
    &\quad = \texttt{Next}\left(X_k^m(t_0-2), p_{\text{TD}}(t_0 -2) \right)
\end{align}

which is the same as obtaining $X_k^*(t_0-2)$ in \eqref{eq:p_tm} and \eqref{eq:app-sample-t-2} 

\hfill $\square$

\subsection{Experimental Verification}
\label{appendix:metrics}

We note that achieving the losslessness described above requires generating draft probabilities such that $p_{\text{DD}}^m(t_0-2) = p_{\text{TD}}(t_0 -2)$. That is, the output distribution of the draft block should be identical to the distribution produced by the same block during true vanilla inference. In some settings, ensuring this is computationally expensive. Concretely, suppose we have 3 blocks $X_1, X_2, X_3$ and we are currently decoding $X_2(t)$ with a draft block $X_2^*(t-1)$. Since $X_2^*$ attends to the entire sequence (excluding $X_2$), but $X_1$ and $X_3$ in turn attend to $X_2$, the draft block is indirectly attending to the current block at a previous timestep $t$. This leakage of information would lead to drift in the output hidden states of the draft block across layers. However, two factors mitigate this deviation. First, $X_1$ has already been fully decoded in semi-autoregressive inference, so its hidden states remain stable during decoding \cite{fastdllmtrainingfreeaccelerationdiffusion}. Second, since $X_3$ consists entirely of masked tokens, both $X_2$ and $X_2^*$ place low attention mass on this block. Due to these factors, the shift in output distribution is minimal. 

In this framework, we may also conclude that when a dLLM is run with blockwise-causal masking, as is done in the SDAR model family \cite{cheng2025sdarsynergisticdiffusionautoregressionparadigm}, the previous hidden states do not attend to future hidden states, which will ensure that the draft and true distribution remain identical since $X_3$ will not attend to $X_2$. Similarly, if a dLLM is run with dual KV caches, before and after the current block \cite{fastdllmtrainingfreeaccelerationdiffusion}, the keys and values of $X_1$ and $X_3$ remain constant, resulting in an exact draft distribution.

We validate our claims by computing the downstream accuracy of the LLaDA, Dream, and SDAR models with and without Spiffy and comparing their resulting metric scores. These results are displayed in Table \ref{tab:main-exp} and demonstrate Spiffy's ability to preserve the model's accuracy. 

We further investigate minor differences in the metric values and attribute them to minor variations in generated outputs due to precision loss caused by floating-point matrix multiplication on CUDA GPUs. This is caused by differing accumulation orders for tensors of differing sizes. In particular, even though each of the $3$ draft blocks used by Spiffy are independent, matrix multiplications with this extended tensor will have a slightly different accumulation order as compared to an input tensor with no draft blocks appended. This leads to minor differences in generations and due to the cumulative nature of errors, may lead to minor deviations in the final generations. For applications where precision is critical, we recommend padding the input tensor to $4096$ to ensure predictable accumulation on CUDA GPUs in reduced precision. We do not adopt this method here as it is prohibitively expensive and minor differences do not affect the output significantly as demonstrated by Table \ref{tab:main-exp}. The PyTorch documentation on numerical accuracy may be referred to for further explanation of floating point error accumulation in such settings.

\section{Extended Experiments}
\label{appendix:open-text-exp}

In Table \ref{tab:open-text-exp}, we use the graphs calibrated on 50 samples of MATH500 and MBPP and evaluate Spiffy on MMLU tasks from medicine, ethics, and law. We see that for diverse evaluation domains, Spiffy's calibration algorithm produces draft graph topologies that effectively capture the dLLM's decoding behavior, resulting in consistent acceleration.

\begin{table*}[h]
    \caption{Spiffy results in consistent \cg{NFE reductions} and \co{TPS speedup} with preserved \cb{accuracy} across tasks from medicine, ethics, and law.}
    \centering
    \small
    \setlength{\tabcolsep}{5pt}
    \begin{tabular}{lccc}
    \toprule
           & MMLU (clinical knowledge) & MMLU (moral scenarios) & MMLU (professional law) \\
    \midrule
    Baseline (LLaDA-8B-Instruct)
        & $\cg{1.00\times}\ (\co{1.00\times}),\ \cb{0.55}$
        & $\cg{1.00\times}\ (\co{1.00\times}),\ \cb{0.35}$
        & $\cg{1.00\times}\ (\co{1.00\times}),\ \cb{0.31}$ \\
    + Dynamic Unmasking
        & $\cg{1.87\times}\ (\co{1.89\times}),\ \cb{0.54}$
        & $\cg{1.62\times}\ (\co{1.60\times}),\ \cb{0.36}$
        & $\cg{1.52\times}\ (\co{1.47\times}),\ \cb{0.32}$ \\
    + Spiffy
        & $\cg{3.18\times}\ (\co{2.38\times}),\ \cb{0.52}$
        & $\cg{2.82\times}\ (\co{2.07\times}),\ \cb{0.38}$
        & $\cg{2.71\times}\ (\co{1.93\times}),\ \cb{0.28}$ \\
    \bottomrule
    \end{tabular}
\label{tab:open-text-exp}
\end{table*}

\section{Overheads of Spiffy}
\label{appendix:overheads}
In this section, we examine the overheads of various stages of Spiffy including drafting, custom attention mask construction, custom position ID construction, and the increase in model inference time due to the extra computation of the draft blocks. We normalize these values to be a percentage of the vanilla model inference time for ease of comparison. This data is collected with LLaDA-8B-Instruct using the MBPP dataset, with the average value reported. All profiling was done using a single 80GB NVIDIA A100 GPU with \texttt{torch==2.6.0} and \texttt{transformers==4.52.4}, with vectorized PyTorch operations and \texttt{torch.nn.scaled\_dot\_product\_attention} with the \texttt{SDPBackend.MATH} kernel.

We see in Table \ref{tab:overheads} that the costs of the various stages of generation with Spiffy are low compared to the model inference time, with drafting in total taking $\sim 6\%$ of the model inference time and draft acceptance taking $\sim 5\%$ of the model inference time. These costs may be reduced via further optimizations to the codebase including additional vectorization and multi-processing for latency-critical scenarios. The model forward pass is $25\%$ more expensive than vanilla inference due to the extra computational requirements and may be mitigated by implementing a custom attention kernel designed for Spiffy's custom attention mask.

\begin{table*}[h]
    \caption{Overheads of Spiffy, normalized to a $\%$ of the model inference time for \# draft blocks $D=3$. $D=0$ corresponds to vanilla inference. All values are normalized to be a percentage of the vanilla model inference time.}
    \centering 
    \begin{tabular}{lcc}
    \toprule
    \multicolumn{3}{c}{\textbf{Overheads as \% of model call time}} \\
    \midrule
    & \multicolumn{2}{c}{Number of draft blocks} \\
    \midrule
     & 0 & 3 \\
    Model &  $100.00$ &  $124.85$  \\
    Draft Graph Construction & $0.00$ & $0.71$\\
    Draft Pruning & $0.00$ & $1.77$ \\
    Draft Sequence Construction & $0.00$ & $0.76$ \\
    Attention Mask Construction & $0.00$ & $0.16$ \\ 
    Position IDs Construction & $0.00$ & $0.05$ \\
    Draft Acceptance & $0.00$ & $6.18$ \\ 
    \bottomrule
    \end{tabular}

    \label{tab:overheads}
\end{table*}

\section{Calibration of Directed Draft Graphs}
\subsection{Intuition}
\label{appendix:calibration-intuition}

In Alg. \ref{alg:calibration}, we first establish a set of graph candidates $G'$ that satisfy our constraints of budget and connectedness: 
\begin{align}
    & \text{A node $q_{kv}$ is \textbf{connected} to a graph $g$ if }  k\in \{1, 2\} \text{ or } parents(q_{kv}) \cap g \neq \varnothing \\
    & \text{A graph $g\in G'$  is connected if it consists of nodes \{$q_{kv}$\} which are connected } \\
    & G' \gets \{g \in G \mid g \text{ is connected AND } |g| = D\}
\end{align}
Intuitively, this means that to be selected, a node must be in the first two levels or descended from a node in the first two levels. The first two levels account for cases in which multi-token decoding is the most conservative. Hence, prioritizing having draft states for such cases is useful. With this constraint, we optimize the following expression: 
\begin{align}
\label{eq:local_count}
G^* = \arg \max_{Q\in G'} \sum_{q\in Q}\left(count(q)\right) 
\end{align}
Each node $q \in Q$ represents a sequence of tokens that has appeared in order. For example, $q=(c_{11}, c_{21}, c_{31})$. $count(q)$ thus represents how often this particular sequence has occurred in the collected calibration data. 

Since we create the full graph $G$ by filtering for the top nodes per level and the top in\_edges per level, we are implicitly introducing parent-child frequency relationship. Moreover, since we may only choose graphs that are connected, we are reinforcing this cross-node connection. For example, $(c_{11}, c_{12})$ or $(c_{11}, c_{31})$ or $(c_{21}, c_{31})$ must have first occurred for $(c_{11}, c_{21}, c_{31})$ to have occurred. So a node is more likely to have occurred if its parents are likely to occur. Thus, using a simple frequency scoring metric, as shown above, suffices to represent the overall score of a candidate draft graph. 

\subsection{Time to Perform Calibration}
\label{appendix:calibration-time}
In our experiments calibrating a draft graph for LLaDA-8B-Instruct on $50$ samples, the time to process a single sample is $\sim 9$ seconds for a total calibration time of $\sim 7$ minutes on a single NVIDIA A100 GPU. The time taken to perform the optimization step in \eqref{eq:local_count} using a brute-force breadth-first-search approach is $\sim 3$ minutes. In sum, calibration is an inexpensive, 10-minute procedure that can lead to high-quality draft graph structures.

\subsection{Calibration Domain}
\label{appendix:calibration-domain}

In Table \ref{tab:calib}, we analyze Spiffy's acceleration when the calibration dataset source is changed from MATH500 and MBPP to ShareGPT. 50 samples total from these datasets are used in either case. We see that Spiffy's calibration algorithm is able to effectively capture the model's underlying decoding dynamics regardless of the calibration domain, resulting in consistent speedups.

\begin{table*}[!htb]
    \caption{Regardless of the calibration domain, Spiffy results in \cg{NFE reductions} and \co{TPS speedup} with preserved \cb{accuracy} across datasets.}
    \centering
    \small
    \setlength{\tabcolsep}{4pt}
    \begin{tabular}{lcccc}
    \toprule
           & GSM8K & HumanEval & MATH500 & MBPP \\
    \midrule
    Baseline (LLaDA-8B-Instruct)
        & $\cg{1.00\times}\ (\co{1.00\times}),\ \cb{0.79}$
        & $\cg{1.00\times}\ (\co{1.00\times}),\ \cb{0.41}$
        & $\cg{1.00\times}\ (\co{1.00\times}),\ \cb{0.34}$
        & $\cg{1.00\times}\ (\co{1.00\times}),\ \cb{0.37}$ \\
    + Dynamic Unmasking
        & $\cg{3.31\times}\ (\co{3.43\times}),\ \cb{0.79}$
        & $\cg{3.53\times}\ (\co{3.13\times}),\ \cb{0.41}$
        & $\cg{2.52\times}\ (\co{2.64\times}),\ \cb{0.35}$
        & $\cg{5.73\times}\ (\co{5.46\times}),\ \cb{0.37}$ \\
    + Spiffy (MATH500-MBPP calib.)
        & $\cg{4.96\times}\ (\co{3.79\times}),\ \cb{0.79}$
        & $\cg{5.85\times}\ (\co{3.59\times}),\ \cb{0.38}$
        & $\cg{4.11\times}\ (\co{3.15\times}),\ \cb{0.35}$
        & $\cg{8.58\times}\ (\co{5.89\times}),\ \cb{0.37}$ \\
    + Spiffy (ShareGPT calib.)
        & $\cg{4.97\times}\ (\co{3.73\times}),\ \cb{0.79}$
        & $\cg{5.85\times}\ (\co{3.49\times}),\ \cb{0.39}$
        & $\cg{4.12\times}\ (\co{3.07\times}),\ \cb{0.35}$
        & $\cg{8.61\times}\ (\co{5.72\times}),\ \cb{0.37}$ \\
    \bottomrule
    \end{tabular}
\label{tab:calib}
\end{table*}

\subsection{Calibration Algorithm Pseudocode}
\label{appendix:calibration-pseudocode}

\definecolor{codegreen}{RGB}{0,128,0}

\lstdefinestyle{pythonstyle}{
    language=Python,
    basicstyle=\ttfamily\small,
    keywordstyle=\color{codegreen}\bfseries,
    commentstyle=\color{codegreen}\itshape,
    stringstyle=\color{codegreen},
    emphstyle=\color{blue},
    emph={calibrate_draft_graph},
    showstringspaces=false,
    breaklines=true,
    columns=flexible,
}

\lstnewenvironment{python}[1][]
    {\lstset{style=pythonstyle,#1}}
    {}
    
\begin{algorithm}[h]
\caption{Draft Graph Calibration Pseudocode}
\label{alg:calibration-pseudo}
\begin{python}
def calibrate_draft_graph(dLLM, dataset, lookahead L, draft_budget D):

    sequences = [record_unmasking_trajectory(dLLM, sample, L)
                 for sample in dataset]
    # Each trajectory = [ (lookahead_depth, { (position_rank(i), vocab_rank(j)) } ) ]

    G = build_candidate_graph(sequences, L)
    # G = graph with L levels
    # Nodes = most frequent unmasked token sets at each level = lookahead depth
    # Eg. level 1 = (1, 1), (2, 1)
    # Edges = most frequent token set transitions
    # Eg. {(1, 1)} -> {(1, 1), (2, 1), (3, 1)}

    G_opt = select_best_subgraph(G, D)
    # G* = argmax_(Q in {size-D connected subgraphs of G}) (sum of node frequencies of Q)

    return G_opt
    # list of D draft nodes, each node is a set of {(pos_rank, vocab_rank)}
\end{python}
\end{algorithm}


\end{document}